\documentclass{article}
\usepackage{graphicx, float} % Required for inserting images
\graphicspath{{images/}}
\usepackage[utf8]{inputenc}
\usepackage{amsmath, amssymb, amsthm}
\usepackage{booktabs}
\usepackage{authblk}

\title{An Enhancement of Cuckoo Search Algorithm for Optimal Earthquake Evacuation Space Allocation  in Intramuros, Manila City}

\author[1]{Marcus Andre Villanueva}
\author[2]{Charles Matthew Ching}
\author[3]{Khatalyn Mata}
\affil[1,2,3]{Department of Computer Science, Pamantasan ng Lungsod ng Maynila, Manila, Philippines}

\date{December 2024}

\begin{document}

\maketitle

\begin{abstract}
    The Cuckoo Search Algorithm (CSA), while effective in solving complex optimization problems, faces limitations in random population initialization and reliance on fixed parameters. Random initialization of the population often results in clustered solutions, resulting in uneven exploration of the search space and hindering effective global optimization. Furthermore, the use of fixed values for discovery rate and step size creates a trade-off between solution accuracy and convergence speed. To address these limitations, an Enhanced Cuckoo Search Algorithm (ECSA) is proposed. This algorithm utilizes the Sobol Sequence to generate a more uniformly distributed initial population and incorporates Cosine Annealing with Warm Restarts to dynamically adjust the parameters. The performance of the algorithms was evaluated on 13 benchmark functions (7 unimodal, 6 multimodal). Statistical analyses were conducted to determine the significance and consistency of the results. The ECSA outperforms the CSA in 11 out of 13 benchmark functions with a mean fitness improvement of 30\% across all functions, achieving 35\% for unimodal functions and 24\% for multimodal functions. The enhanced algorithm demonstrated increased convergence efficiency, indicating its superiority to the CSA in solving a variety of optimization problems. The ECSA is subsequently applied to optimize earthquake evacuation space allocation in Intramuros, Manila.
\end{abstract}


\section{Introduction}

The Philippines, located on the Pacific Ring of Fire, is highly vulnerable to a wide range of natural disasters, such as earthquakes. One of the most significant seismic threats of the country is the West Valley Fault, which traverses the densely populated region of Metro Manila and neighboring provinces. Given the potential for catastrophic damage, efficient disaster management is crucial in mitigating damages caused by natural disasters. An essential part of disaster management is the allocation of evacuation spaces as it prioritizes short-term safety during critical hours to reduce casualties and aid affected populations. This is known as a Location Allocation (LA) problem.

Recent literatures have approached LA problems using optimization algorithms inspired by nature observation, which are known as metaheuristics algorithms. For instance, Yin et al. \cite{10.3389/fpubh.2022.1098675} proposed a quantum Genetic Algorithm to solve emergency shelter allocation planning for large-scale evacuation. Moreover, Bae et al. \cite{bae2019shelter} utilized the Genetic Algorithm to find the optimal solution in shelter location-allocation for tsunamis in Korea using floating point population to reflect reality in emergency situations. The utilization and modification of the Particle Swarm Optimization algorithm was also explored by Hu et al. \cite{hu2012modified} on the allocation problem of earthquake emergency shelters. 

These relevant studies have proven that metaheuristic algorithms are an effective approach when solving LA problems. One of the well-known metaheuristic algorithms used extensively is the Cuckoo Search Algorithm (CSA) \cite{yang2009cuckoo}. 

The CSA is inspired by the behavior of a specific species of cuckoo birds called Ani and Guira, which are brood parasites that do not build their own nest, but rather, intrude and lay their eggs into a host’s nest. The cuckoos often choose nests where the host bird recently laid its own eggs. In general, cuckoo eggs hatch slightly earlier than host eggs. Some host birds can engage in direct conflict with intruding cuckoos. If a host bird discovers an alien egg, it will either throw it or abandon its nest and build a new nest elsewhere \cite{santos2012cuckoo}. Alternatively, if the first cuckoo chick is hatched, its initial instinct is to evict the host eggs by blindly pushing them out of the nest. This increases the cuckoo chick’s share of food and aims to secure a larger portion of the host bird’s food resources \cite{valian2011improved}.

While the CSA has proven its effectiveness, it has several limitations that need to be further studied. Among its limitations is the initialization process of CSA. Since the population of cuckoos is initialized randomly, the solutions may not be evenly distributed across the search space. Some areas of the search space might be densely populated, while others may be sparsely explored or not explored at all. According to Sun \cite{sun2024application}, the random distribution policy for initializing the population can lead to clustering in specific areas, which hinders global optimization. 

Another limitation are the fixed values for the discovery rate ($P_a$) and step size ($a$) of CSA. The algorithm uses fixed values for both discovery rate and step size . These values are defined during the initialization step and cannot be changed during new generations. Valian et al. \cite{valian2011improved}, in their study “Improved Cuckoo Search Algorithm for Global Optimization,” noted that using fixed parameter values can negatively impact the number of iterations needed to find an optimal solution, ultimately reducing the algorithm’s efficiency. 

In this study, the researchers aim to enhance the CSA by improving the limitations of the algorithm in initial population, and exploration-exploitation balance. The researchers then aims to apply the algorithm for optimal earthquake evacuation space allocation in Intramuros, Manila City.

\section{Literature Review}
Some studies have been done that explored the different initialization methods for metaheuristic algorithms, including quasi-random initialization, chaotic systems, anti-symmetric learning methods, and Latin hypercube sampling \cite{li2020influence}. Despite the methods being able to enhance the performance of metaheuristic algorithms such as Particle Swarm Optimization (PSO) and Genetic Algorithms (GA), they have significant drawbacks. Quasi-random initialization struggles with the curse of dimensionality \cite{maaranen2004quasi}. Chaos-based approaches produce random sequences using a few chaotic maps and parameters, but they are highly sensitive to initial conditions in certain situations \cite{dos2008use}. The anti-symmetric learning method demands twice the population size to select solutions for the next generation, thereby doubling the computational cost. Latin hypercube sampling is effective in low-dimensional spaces but can perform poorly in higher-dimensional ones.

Meng et al. \cite{meng2019multi} proposed a population strategy based on constraint transformation and the individual constraints and group constraints technique (ICGC) to improve the population initialization for increased search efficiency. This was proposed to overcome the shortcoming of the multi-objective cuckoo search (MOCS), which was developed by Yang \& Deb \cite{yang2013multiobjective} to the multi-objective optimization problem. The improved algorithm was applied to the multi-objective hydropower station optimal operation (MOHSOO) model of Xialongdi and Xixiayuan cascade hydropower stations in the Yellow River. The results show that the population strategy can enhance search efficiency by constraining the initial solution within a defined range, thereby reducing the search space and enhancing the quality of the initial feasible solution. However, the study mentioned that population initialization strategy using ICGC may be more suitable for short-term hydropower station operation, as monthly water level variations are significant and may not fully utilize ICGC's potential. Thus, exploring more effective improvement strategies and applying them to more complex models is crucial for better performance.

According to Valian et al. \cite{valian2011improved} from their study, “Improved Cuckoo Search Algorithm for Global Optimization”, a key drawback of using fixed values is that it affects the number of iterations required to find an optimal solution, which may lead to decreasing the efficiency of the algorithm. To enhance the performance and convergence rate of the CS algorithm and address the limitations associated with fixed parameter values, an adaptive method for the parameters was proposed. The enhanced CS algorithm was called Improved Cuckoo Search (ICS) algorithm in the study. The primary distinction between the ICS and CS algorithms lies in the way $P_a$ and $a$ are adjusted. The values of $P_a$ and $a$ are dynamically changed with the numbers of generations, wherein the values of $P_a$ and $a$ are large to enhance the diversity of solution vectors and are gradually decreased in the final generations for better fine-tuning of solution vectors. According to the simulation results, the ICS algorithm demonstrated its superiority over the standard CS algorithm in terms of both accuracy and convergence rate in several benchmark problems. The study showed that for varying population sizes (N = 10, 30, and 50) and iteration counts (1000, 3000, and 5000), dynamically adjusting $P_a$ and $a$ had a significant impact on the algorithm's performance. The results indicated that reducing the minimum value of $P_a$ without altering $a$ led to better results in most cases, especially for test functions with high decision variables, while increasing the maximum value of $P_a$ also enhanced performance. Although reducing the minimum value of $\alpha$ had little impact, increasing $\alpha$ excessively could degrade performance. Based on the tests, optimal values for the parameters were identified as approximately  
$P_{a\text{min}} = 0.005, \quad P_{a\text{max}} = 1, \quad a_{\text{min}} = 0.05, \quad a_{\text{max}} = 0.5.$

A study entitled “Improved cuckoo search algorithm and its application to permutation flow shop scheduling problem” by Zhang et al. \cite{zhang2020improved}, developed a self-adaptive step length CS algorithm based on a dynamic balance factor, referred to as the dynamic cuckoo search (DCS) algorithm. According to Zhang et al. \cite{zhang2020improved}, in the standard CS algorithm, while a smaller step size improves local search accuracy, it slows down convergence and increases the risk of getting trapped in local optima. Conversely, a larger step size enhances global search and accelerates convergence but may cause the algorithm to skip the optimal value, resulting in oscillation around it or even population loss. The randomness of the Levy flight does not guarantee that the search process converges quickly and steadily towards the optimal solution. Zhang et al. \cite{zhang2020improved}, mentioned that a better approach is to integrate both global and local search capabilities, striking a balance between fast global exploration and precise local optimization to improve overall algorithm performance. The DCS algorithm introduces two key parameters, the iteration number ratio parameter and the adaptability ratio parameter, which are controlled by a dynamic balance factor ($\vartheta$). The dynamic balance factor ($\vartheta$) is introduced and used to adjust the weight number of iteration number ratio and adaptability ratio. The DCS algorithm adjusts these parameters throughout the optimization process to achieve a more flexible and adaptive search, closely mimicking natural evolution. Additionally, the self-adaptive step is realized through the dynamic adjustment of the parameter $\beta$ in the standard CS algorithm, enhancing the algorithm’s ability to explore and exploit the search space. The effectiveness of the DCS algorithm was verified using six standard test functions (Ackley, Griewank, Rastrigin, Rosenbrock, Schwefel, and Sphere) and demonstrated superior performance in solving the permutation flow shop scheduling problem (PFSP), particularly when tested against eight operators Carl-Car8 of Car benchmark class.

A multi-strategy adaptive cuckoo search algorithm was proposed to further enhance a self-adaptive variant of the cuckoo search algorithm which aims to balance the exploration (global search) and exploitation (local search) capability of the algorithm as well as address the limitations of the variant. In this study, 5 different search strategies were proposed which are fairly controlled by a certain probability. The different search strategies used by the algorithm comprises of the following: (1) Maintaining the original global and local search of the cs algorithm, (2 \& 3) Selection of 2 or 4 individuals for location information sharing, (4) A search strategy with a learning matrix to record solutions, and (5) Increased randomness when searching and made the whole population jump out of the local optimal region by the nonlinear inertia coefficient. The multi-strategy adaptive cuckoo search algorithm yielded positive results by achieving the best result on 17 out of 28 optimization benchmark functions \cite{gao2021adaptive}.

An experimental study was conducted by Salgotra et al. \cite{salgotra2018new} wherein multiple variants of the cuckoo search algorithm that enhances both global and local search were proposed. The three versions of the algorithm are as follows: (1) in version 1, the exploration enhancement is based on the search equations of the Grey Wolf Optimizer while the exploitation enhancement utilized two different searching strategies to evaluate the final solution, (2) in version 2, the exploration and exploitation enhancement focused on two-fold division in population using a pair of equation on the divided populations to generate new solutions, (3) lastly in version 3, a Cauchy based mutation operator was added to the explorative capabilities of CS and a four-fold division of population in the global search space was also added to balance exploration and exploitation using different search equations per part. Out of the three proposed versions, the Cuckoo search version 1 outperformed state-of-the-art algorithms in the experiments which used Cauchy based exploration, and better search of the search space.

\section{Methods}
The standard CSA follows three idealized laws based from the specific behavior of the cuckoos: (1) Each cuckoo lays one egg at a time and disposes the egg at a randomly chosen host nest; (2) The best nests with excellent quality of eggs will carry over to the next generation; (3) The number of host nest is fixed and the probability of parasitic egg discovery is $P_a$ [0,1] \cite{roy2013cuckoo}. The last assumption can be approximated by a fraction of $P_a$ of the $n$ nests, having new random solutions.

\begin{figure}[htbp]
    \centering
    \includegraphics[width=4in]{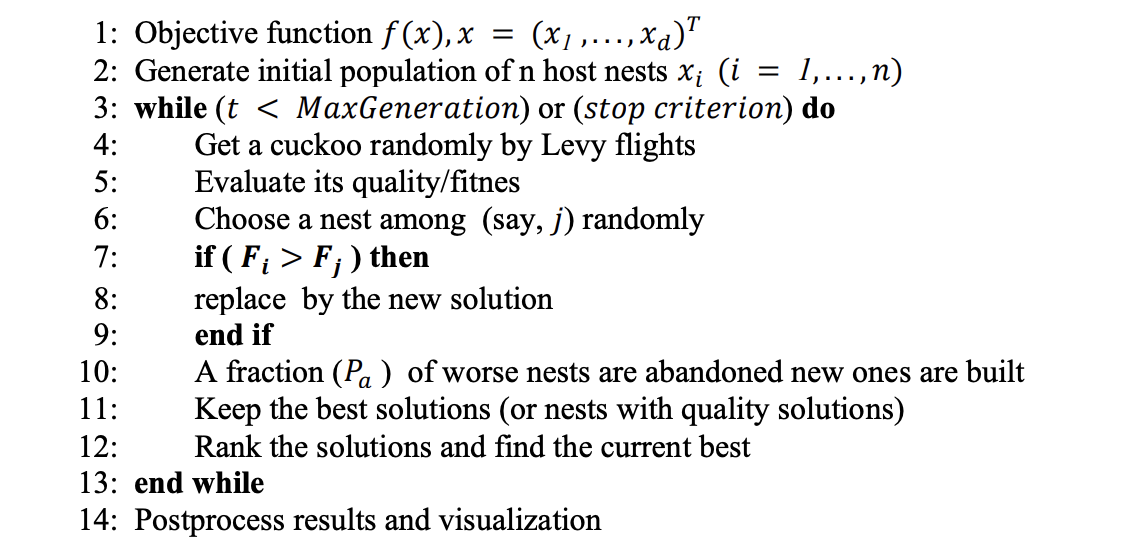}
    \caption{Pseudocode of the Cuckoo Search Algorithm}
    \label{fig:csa pseudocode}
\end{figure}

With the limitations of the standard CSA, such as random population initialization (line 2 of the pseudocode) and the use of fixed parameters (line 10 of the pseudocode), this study proposes an Enhanced Cuckoo Search Algorithm (ECSA) incorporating a well-distributed population initialization method and self-adaptive parameters to improve convergence efficiency.

Sobol Sequence is implemented at the initialization stage to generate a more uniformly distributed initial population for improved global search. The Sobol Sequence was invented in the USSR in 1967 by Russian mathematician Ilya M. Sobol. Its computation was massively improved in 1979 by Antonov and Saleev, which resulted in a better generation of numbers with only a few low-level bit-wise operations \cite{sirsant2022improved}. Due to its lower discrepancy nature, Sobol numbers result in a faster convergence and more stable estimates of values. The Sobol sequence is based on the concept of direction numbers and the bitwise exclusive OR (XOR) operation to generate points that are uniformly distributed across a multi-dimensional space. This makes it highly effective for applications requiring high-quality sampling. Sobol Sequence is implemented on the line 2 of the pseudocode.

\begin{figure}[htbp]
    \centering
    \includegraphics[width=4.5in]{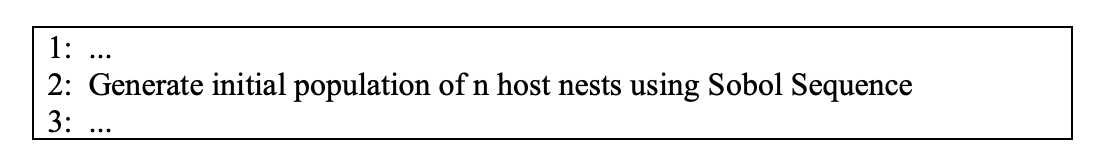}
    \caption{Sobol Sequence implementation}
    \label{fig:sobol sequence implementation}
\end{figure}

Cosine Annealing with Warm Restarts is used to dynamically adjust the discovery rate and step size on the current iteration of the algorithm, taking advantage of a cosine-shaped schedule for an improved convergence rate. Cosine Annealing is a learning rate schedule, which is a common technique used in training deep learning models. Cosine Annealing starts with a high learning rate that gradually decreases to a minimum value and once reaching the minimum value, the learning rate resets by rapidly increasing the learning rate to the maximum which is done in cycles over the whole training period. Cosine Annealing with Warm Restarts is added after line 12 of the pseudocode.

\begin{figure}[htbp]
    \centering
    \includegraphics[width=4.5in]{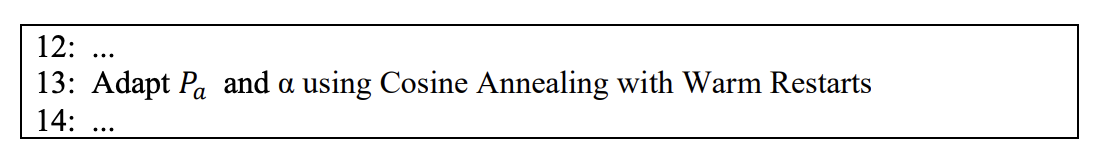}
    \caption{Cosine Annealing with Warm Restarts implementation}
    \label{fig:cosine annealing implementation}
\end{figure}

\begin{figure}[H]
    \centering
    \includegraphics[width=4.5in]{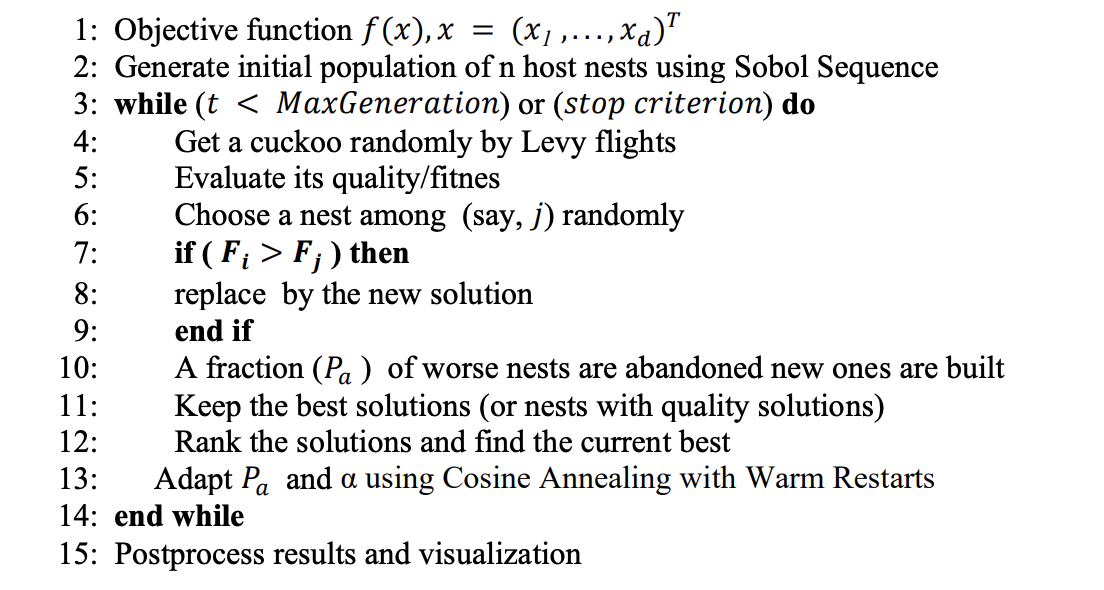}
    \caption{Pseudocode of Enhanced Cuckoo Search Algorithm}
    \label{fig:ecsa pseudocode}
\end{figure}

Statistical analysis was incorporated to compare and evaluate the performance of the CSA and the ECSA against thirteen benchmark functions. The researchers utilized Sphere, Schwefel’s 2.22, Schwefel’s 1.20, Schwefel’s 2.21, Rosenbrock’s, Step, and Quartic Noise benchmark functions to test the algorithms in a unimodal optimization problem. While Rastrigin, Ackley, Griewank, Schwefel, Generalized Penalized 1, and Generalized Penalized 2 were utilized to test the algorithms in a multimodal optimization problem. These functions aim to assess the local and global searchability of the algorithms and validate the enhancements made to the existing algorithm.

Subsequently, the performance analysis was extended to the application of this study, which is a discrete optimization task. The researchers utilized a developed LA model derived from Cooper’s General LA model \cite{azarmand2009location}. The function minimizes the distance between the evacuation areas and district blocks. A lower fitness score indicates a more optimally allocated solution of the evacuation areas.

\begin{figure}[htbp]
    \centering
    \begin{equation}
        \textit{Fitness}(x) = \sum_{i=1}^{m} \sum_{j=1}^{n} d(x_i, x_j)
    \end{equation}
    \caption{Where \(\mathbf{x}\) represents a solution and \( d(x_i, x_j) \) represents the distance between block \( j \) and evacuation area \( i \).}
\end{figure}

The researchers gathered data based on the land use map of Intramuros in 2015 and the official database of the Intramuros Administration for plazas, parks, and open spaces. The researchers utilized QGIS, an open-source geographic information system tool, to map the district blocks using a polygon feature and candidate evacuation areas using a point feature. The coordinates of the district blocks were identified by extracting the centroid of the polygon features.

\section{Results}

The researchers ran 30 individual simulations of the algorithms against the thirteen optimization benchmark functions. The obtained function evaluation values (FEVs) were used to calculate the mean, standard deviation, and p-value (Wilcoxon rank sum test). Shown below are the parameters used in this experiment.

\begin{table}[htbp]
    \centering
    \begin{tabular}{lcccccc}
        \toprule
        & \textbf{Dimension} & \textbf{Population} & \textbf{Iterations} & $P_a$ & $\alpha$ \\
        \midrule
        CSA  & 15 & 50 & 500 & 0.25 & 0.01 \\
        ECSA & 15 & 50 & 500 & [0.5,0.25] & [0.01,0.05] \\
        \bottomrule
    \end{tabular}
    \caption{Parameters settings of CSA and ECSA}
    \label{tab:parameters}
\end{table}

\begin{table}[htbp]
    \centering
    \begin{tabular}{lcc|cc}
        \toprule
        & \multicolumn{2}{c}{\textbf{CSA}} & \multicolumn{2}{c}{\textbf{ECSA}} \\ 
        \cmidrule(lr){2-3} \cmidrule(lr){4-5}
        & \textit{mean} & \textit{std} & \textit{mean} & \textit{std} \\ 
        \midrule
        \textit{F1}  & 2.03e-107  & 1.08e-106  & \textbf{1.21e-132}  & \textbf{6.29e-132}  \\ 
        \textit{F2}  & 4.63e-59  & 1.45e-58  & \textbf{5.55e-71}  & 1.98e-70  \\ 
        \textit{F3}  & 1.24e-86  & 6.40e-86  & \textbf{7.12e-112}  & 3.83e-111  \\ 
        \textit{F4}  & 1.46e-52  & 5.21e-52  & \textbf{8.60e-65}  & 3.59e-64  \\ 
        \textit{F5}  & \textbf{13.552799} & \textbf{0.568044} & 13.333362 & 0.661025 \\ 
        \textit{F6}  & 1.825075  & 0.555273  & \textbf{1.545722}  & \textbf{0.470394}  \\ 
        \textit{F7}  & 0.000523  & 0.000420  & \textbf{0.000306}  & \textbf{0.000196}  \\ 
        \bottomrule
    \end{tabular}
    \caption{Results of the unimodal experiment}
    \label{tab:unimodal}
\end{table}

Table 2 shows the recorded mean and standard deviation of unimodal benchmark functions F1 – F7. Based on these results, ECSA outperformed CSA on majority of the selected unimodal benchmark functions. The algorithm achieved lower mean and standard deviation on F1 – F4 and F7 which suggests that it was able to consistently reach values closer to the minimum value. In contrast, the algorithm achieved mixed results on benchmark functions F5 and F6. ECSA was able to achieve slightly less mean and higher standard deviation than CSA on F5 which suggests that while there is slight variability, ECSA was able to find better solutions. Furthermore, ECSA was able to achieve slightly lower mean and standard deviation on F6. The mixed results from F5 and F6 suggest that while the algorithm improved in reaching values closer to the minimum value, the improvement is not significant when solving similar problem types.

\begin{table}[htbp]
    \centering
    \begin{tabular}{lcc|cc}
        \toprule
        & \multicolumn{2}{c}{\textbf{CSA}} & \multicolumn{2}{c}{\textbf{ECSA}} \\ 
        \cmidrule(lr){2-3} \cmidrule(lr){4-5}
        & \textit{mean} & \textit{std} & \textit{mean} & \textit{std} \\ 
        \midrule
        \textit{F8}  & 0.036305  & 0.195407  & \textbf{7.27e-07}  & 3.46e-06  \\ 
        \textit{F9}  & 1.36e-09  & 5.08e-09  & \textbf{7.57e-12}  & 2.55e-11  \\ 
        \textit{F10} & 0.021790  & 0.082574  & \textbf{0.000571}  & 0.002892  \\ 
        \textit{F11} & -3293.31  & 499.35    & \textbf{-3776.53}  & 422.29    \\ 
        \textit{F12} & 0.648578  & 0.381149  & \textbf{0.426125}  & 0.291521  \\ 
        \textit{F13} & 12.364474 & 1.418755  & \textbf{11.407482} & 1.416843  \\ 
        \bottomrule
    \end{tabular}
    \caption{Results of the multimodal experiment}
    \label{tab:multimodal}
\end{table}

Table 3 shows the recorded mean and standard deviation of multimodal benchmark functions F8 – F10. Based on these results, ECSA outperforms CSA on all the selected multimodal benchmark functions. ECSA achieved a lower mean and standard deviation than CSA which suggests that the algorithm was able to find solutions closer to the minimum value at a much faster rate. Furthermore, the low standard deviation also indicates that ECSA was able to achieve this result consistently. The comparison made on the optimization results for F8 – F13 demonstrates that ECSA is also superior to CSA in solving multimodal type of problems.

\begin{table}[htbp]
    \centering
    \begin{tabular}{cc|cc}
        \toprule
        \multicolumn{2}{c|}{\textbf{CSA vs ECSA}} & \multicolumn{2}{c}{\phantom{}} \\ 
        \midrule
        \textit{F1} & 1.86e-09 & \textit{F8} & 3.79e-06 \\
        \textit{F2} & 1.86e-09 & \textit{F9} & 1.30e-07 \\
        \textit{F3} & 1.86e-09 & \textit{F10} & 4.40e-05 \\
        \textit{F4} & 1.86e-09 & \textit{F11} & 0.000418 \\
        \textit{F5} & 0.212 & \textit{F12} & 0.003222 \\
        \textit{F6} & 0.076 & \textit{F13} & 0.015 \\
        \textit{F7} & 0.027 & & \\
        \bottomrule
    \end{tabular}
    \caption{The p-values of the Wilcoxon rank sum test}
    \label{tab:wilcoxon}
\end{table}

Table 4 shows the recorded p-value of the statistical test, carried out with a 5\% confidence level. The null hypothesis states that if the p-value >= 0.05, the algorithms are comparable while the alternative hypothesis states that if p-value < 0.05, the algorithms are significantly different. In 11 out of 13 benchmark functions, it shows statistically significant differences between the performance of CSA and ECSA in reaching the optimum solution, indicating that ECSA is superior to CSA.

The researchers were able to gather 50 district blocks and 11 candidate evacuation areas. To solve the model, the researchers discretized the CSA and the ECSA using one-hot encoding to represent a solution using a binary vector. ECSA achieved a mean and standard deviation of 1.41e-4 and 1.62e-4 respectively while CSA achieved a mean and standard deviation of 2.11e-4 and 1.73e-4 respectively. Additionally, the discrete optimization results obtained a p-value of 0.129, indicating that, statistically, the performance of the algorithms is comparable in a discrete space.

\begin{figure}[htbp]
    \centering
    \includegraphics[width=4.5in]{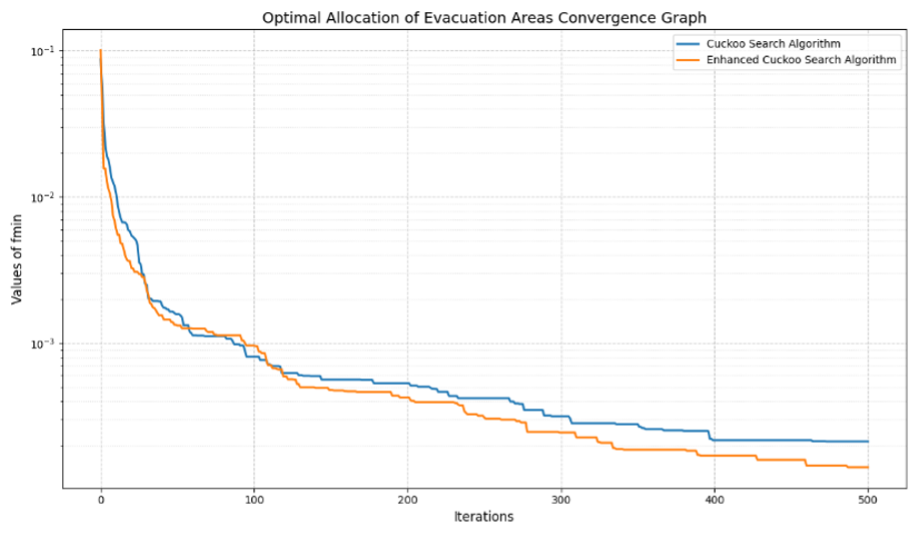}
    \caption{Convergence Graph of discrete optimization results}
    \label{fig:dsa convergence graph}
\end{figure}

However, it can be observed that the performance trend is in favor of ECSA. Figure 6 illustrates the convergence graph of the discrete optimization results, highlighting the performance of ECSA compared to CSA. The convergence graph demonstrates that ECSA converges faster and consistently, achieving lower fitness values throughout the optimization process. The results from this experiment indicate that ECSA is a more accurate, reliable, and a faster algorithm than CSA for solving discrete optimization problems.

\section{Discussion}

Addressing the limitations of the CSA in initial population distribution and fixed parameters was achieved by incorporating Sobol Sequence for a well-distributed population initialization and utilizing Cosine Annealing with Warm Restarts to enable dynamic adjustments to the discovery rate and step size. The incorporation of the Sobol Sequence ensured a more uniform exploration of the search space and mitigated the clustering issues associated with random initialization, resulting in faster and more reliable convergence. Similarly, the use of Cosine Annealing with Warm Restarts facilitated a better balance between exploration and exploitation by dynamically adjusting parameters throughout the optimization process.

Through rigorous evaluation against thirteen benchmark functions, the ECSA outperformed the standard CSA in both unimodal and multimodal optimization problems, validating its robustness and adaptability. The results from benchmark testing highlight that the proposed algorithm achieves superior local and global search performance, addressing both convergence speed and solution accuracy.

Additionally, the application of the ECSA to the discrete optimization task of earthquake evacuation space allocation in Intramuros demonstrated its practical relevance. The algorithm effectively minimized the distance between district blocks and evacuation areas, providing a more optimal allocation solution compared to the standard CSA.

In summary, the ECSA represents a novel approach in addressing the recurring limitations of the CSA in initial population distribution and its fixed parameters. The results show improvement in convergence efficiency, solution accuracy, and broad applicability of the ECSA in optimization tasks, even extending to discrete optimization. However, limitation of the trasferability of the enhancements to a discrete setting may influence the validity of the performance results of the ECSA in earthquake evacuation space allocation as shown in the findings of this study.

\section{Conclusion}
This study successfully addresses the key limitations in the CSA regarding the uneven exploration of the search space caused by the clustered solutions of the random initialization process, and the trade-off between solution accuracy and convergence speed caused by the fixed values for discovery rate and step size. Through the introduction of the Sobol Sequence for initial population distribution, the ECSA overcomes the clustering issue observed in the algorithm, enabling a more thorough exploration of the search space. Additionally, the implementation of Cosine Annealing with Warm Restarts provides a dynamic, self-adaptive mechanism for adjusting the discovery rate and step size, balancing solution accuracy with convergence speed. Performance testing on 13 benchmark functions validates these enhancements and the study’s objectives, providing a more efficient algorithm with broad applicability in optimization tasks. Furthermore, the performance of the ECSA in finding the optimal allocation of earthquake evacuation areas shows its adaptability in discrete optimization tasks. Although the findings were statistically insignificant, the performance trend of the algorithm in the convergence graphs suggests that ECSA generally outperforms CSA in discrete spaces.

While this study applied ECSA to earthquake evacuation space allocation, future research could explore its potential in other critical domains, such as supply chain management, renewable energy resource allocation, and healthcare logistics. These applications would validate ECSA’s adaptability and effectiveness in solving optimization problems. Additionally, exploring other search mechanisms may further enhance ECSA. Incorporating a more sophisticated search mechanism can lessen the recklessness in searching the solution space, which could further improve the convergence speed and better escape local optima in multimodal problems.

\section{Limitation}
A key limitation observed in this study is the extension of ECSA in solving a discrete optimization problem. The performance evaluation of this study were conducted in a continuous space, which may limit the transferability of the enhancements to a discrete setting. Specifically, the results showed that the improvements were statistically insignificant when applied to a discrete LA problem. This observation suggests that future studies may focus on tailoring ECSA to better solve discrete optimization problems, which is beyond the scope of this study.

\bibliographystyle{IEEEtran}
\bibliography{references.bib}

\end{document}